\documentclass[12pt]{article}
\usepackage{array}
\usepackage{epsfig}
\usepackage{epstopdf}
\usepackage{rotating}
\usepackage{amsmath}
\usepackage{longtable}
\usepackage{graphicx}
\usepackage{natbib}         
\usepackage{natbib}         

\begin{document}

\title{How Does Latent Semantic Analysis Work? A Visualisation Approach}
\author{Jan Koeman (jpk39@uclive.ac.nz), \\ William Rea (bill.rea@canterbury.ac.nz, corresponding
author) \\Department of Economics and Finance, \\ University of Canterbury,\\ Christchurch, \\
New Zealand}

\maketitle

\begin{abstract}
By using a small example, an analogy to photographic
compression, and a simple visualization using heatmaps,
we show that latent semantic analysis (LSA) is able to
extract what appears to be semantic meaning of
words from a set of documents by blurring the distinctions
between the words. 
\end{abstract}

\begin{description}
\item[Keywords: ] Latent Semantic Analysis, Singular Value Decomposition, 
Artificial Intelligence, Visualization
\end{description}


\section{Introduction}\label{sec:intro}

Latent Semantic Analysis (LSA) was patented in 1988  (US Patent 4,839,853) 
and is a widely used technique in natural language processing for analyzing 
relationships between a set of documents and the words they contain.
The literature on LSA is extensive, see, for example, the book-length
collection \cite{Landauer2011} and the many references therein. Among this
literature are a number of excellent decriptions of the mathematics of LSA
such as \cite{Deerwater1990}, \cite{Berry1995} and \cite{Martin2011}.

Despite the 
existence of these excellent mathematical accounts of
how LSA works, it is not widely 
understood.  For example, recently \cite{Tunkelang2008}, discussed
three alternative hypothesis but reached no conclusions.
We suspect the reason for this is that to understand how LSA
works, substantial knowledge of
linear algebra and matrix computations is required
in general, and of singular value decomposition (SVD) 
in particular, on the level provided in
such textbooks as
\citet[][Sec. 5.12]{Meyer2000}, \citet[][Ch. 4]{Watkins2000} or 
\citet[][Sect. 7.7]{Gentle2007} among
many others. To those without such knowledge the discussion
of the use of singular values or eigenvalues 
(for details of the exact nature of the relationship
between singular values and eigenvalues see \cite{Meyer2000} p.\ 555)
and their corresponding
eigenvectors in LSA is simply incomprehensible.
It is our hope that by using a visualisation approach
in this short note the understanding of how LSA works (at least
at the intuitive level) will become
available to a much wider audience, particularly to users of LSA with
minimal mathematical backgrounds.

In this note we have used the words eigenvalue and eigenector, understanding
that they carry little or no meaning to a non-mathematical reader, simply 
because they are standard terms. 

The remainder of the note only contains two sections; 
Section (\ref{sec:example}) presents an example while
Section (\ref{sec:conclude}) contains our conclusions.

\section{Example}\label{sec:example}

\begin{table}[ht]
\begin{center}
\begin{tabular}{ll}
Document & Title \\
\hline
c1:& {\em Human} machine {\em interface} for ABC {\em computer} applications \\
c2:& A {\em survey} of {\em user} opinion of {\em computer system response times} \\
c3:& The {\em EPS} user interface management {\em system} \\
c4:& {\em System} and {\em human system} engineering testing of {\em EPS} \\
c5:& Relation of {\em user} perceived {\em response time} to error measurement \\
   & \\
m1:& The generation of random, binary, ordered {\em trees} \\
m2:& The intersection {\em graph} of paths in {\em trees} \\
m3:& {\em Graph minors} IV: Width of {\em trees} and well-quasi-ordering \\
m4 & {\em Graph minors}: A {\em survey}\\
   &  \\
\end{tabular}
\begin{tabular}{l|rrrrrrrrrr}
\hline
Document & c1& c2& c3& c4& c5& m1& m2& m3& m4 \\
Word      &   &   &   &   &   &   &   &   &  \\
\hline
human            & 1&  0&  0&  1&  0&  0&  0&  0&  0 \\
interface        & 1&  0&  1&  0&  0&  0&  0&  0&  0 \\
computer         & 1&  1&  0&  0&  0&  0&  0&  0&  0 \\
user             & 0&  1&  1&  0&  1&  0&  0&  0&  0 \\
system           & 0&  1&  1&  2&  0&  0&  0&  0&  0 \\
response         & 0&  1&  0&  0&  1&  0&  0&  0&  0 \\
time             & 0&  1&  0&  0&  1&  0&  0&  0&  0 \\
EPS              & 0&  0&  1&  1&  0&  0&  0&  0&  0 \\
survey           & 0&  1&  0&  0&  0&  0&  0&  0&  1 \\
trees            & 0&  0&  0&  0&  0&  1&  1&  1&  0 \\
graph            & 0&  0&  0&  0&  0&  0&  1&  1&  1 \\
minors           & 0&  0&  0&  0&  0&  0&  0&  1&  1 \\
\hline
\end{tabular}
\end{center}
\caption{The original document titles and word-by-content matrix 
from \cite{Landauer1998} Figure (1)
of the
titles of nine documents. The words selected for the matrix
are words which occur in at least two document titles.
The words chosen are in italics above. The subject matter of documents c1-c5 is 
human-computer interaction, documents m1-m4 are on mathematical
graph theory.}\label{tab:landauerf1}
\end{table}

We begin by presenting an example of the application of
SVD to image processing and compression. 

\begin{figure}[ht]
  \centering
  \includegraphics[width=7cm]{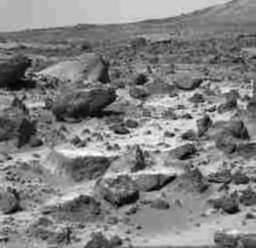}
  \caption{This picture is a 
grayscale
photograph from the Mars Pathfinder mission 
(http://nssdc.gsfc.nasa.gov/planetary/mesur.html) taken on 4 July, 1997.}
  \label{fig:yogi1}
\end{figure}

\begin{figure}[ht]
  \centering
  \includegraphics[width=7cm]{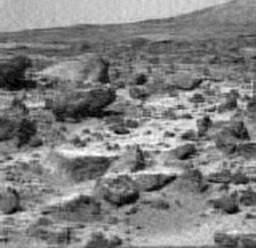}
  \caption{This picture is a reconstruction of the photograph in
Figure (\ref{fig:yogi1}) using the first 36 eigenvectors.
}
  \label{fig:yogi2}
\end{figure}

\begin{figure}[ht]
  \centering
  \includegraphics[width=7cm]{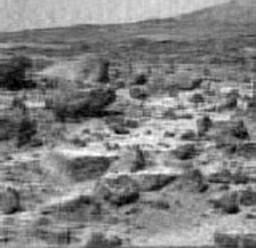}
  \caption{This picture is a reconstruction of the photograph in
Figure (\ref{fig:yogi1}) using the first 25 eigenvectors.
}
  \label{fig:yogi3}
\end{figure}

If
a photograph is subject to an SVD and then the photograph recreated 
from a subset of the eigenvalues 
and corresponding eigenvectors we obtain an approximation to the 
original photograph. 
An example can be seen in Figures (\ref{fig:yogi1}) through (\ref{fig:yogi3}).
Figure (\ref{fig:yogi1}) is the original greyscale
photograph of the Martian landscape taken by the Mars Pathfinder lander
\citep{NASA1997}.  Figures (\ref{fig:yogi2}) and (\ref{fig:yogi3})
are approximations to the original photograph created by
doing an SVD on the photograph's matrix of grayscale values
and recreating it using the
first 36 and 25 eigenvectors respectively. 
The major features 
of the original photograph can been seen in both Figures (\ref{fig:yogi2}) 
and (\ref{fig:yogi3}),
but much of the fine detail has been lost and obviously more
detail has been lost from Figure (\ref{fig:yogi3}) 
than from Figure (\ref{fig:yogi2}). From
the perspective of the human eye the 
 photographs in Figures (\ref{fig:yogi2}) and (\ref{fig:yogi3}) 
are slightly blurry compared to the original.
The degree of blur depends on how few eigenvectors were selected to
recreate the photograph.

We now turn to LSA. 
Table (\ref{tab:landauerf1}) contains the data used by
\cite{Landauer1998} to illustrate LSA. 
\cite{Landauer1998} constructed an example in which there are two
distinct concepts, human-computer interaction (documents c1-c5)
and graph theory (documents m1-m4), which
share only a single common word ``survey''. Frequently
occurring words like
``and'', ``of'' and ``the'' are routinely omitted
because they tend to obfuscate an LSA. 
\cite{Landauer1998} chose words
which appear in at least two titles of their small corpus
for inclusion in the LSA. 
As can be seen from Table (\ref{tab:landauerf1}) the word-document
matrix is quite sparse. 

\begin{figure}[ht]
\begin{center}
\makebox[\textwidth][c]{\includegraphics[width=1.4\textwidth]{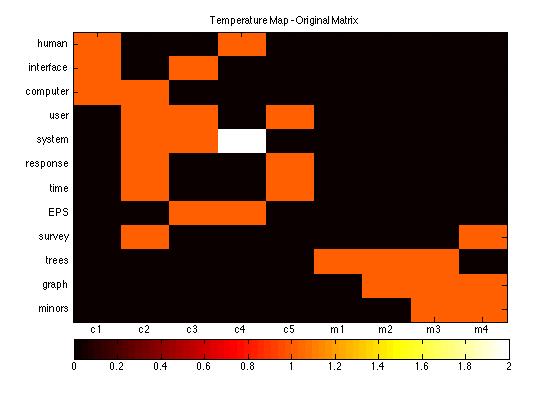}}
\end{center}
  \caption{A heatmap of the full word-document matrix in Table 
(\ref{tab:landauerf1}).} \label{fig:heat1}
\end{figure}

\begin{figure}[ht]
\begin{center}
\makebox[\textwidth][c]{\includegraphics[width=1.4\textwidth]{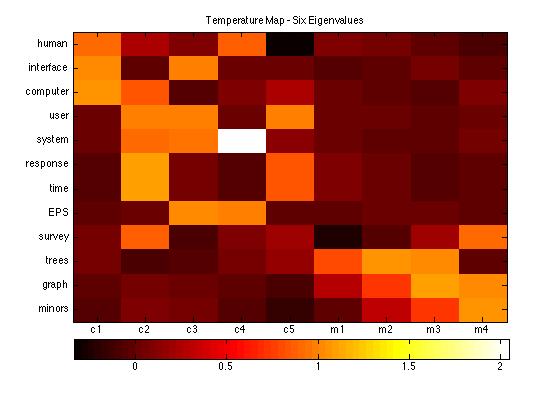}}
\end{center}
  \caption{A heatmap of an approximation to the
word-document matrix in Table 
(\ref{tab:landauerf1}) using the six largest eigenvectors.} \label{fig:heat2}
\end{figure}

\begin{figure}[ht]
\begin{center}
\makebox[\textwidth][c]{\includegraphics[width=1.4\textwidth]{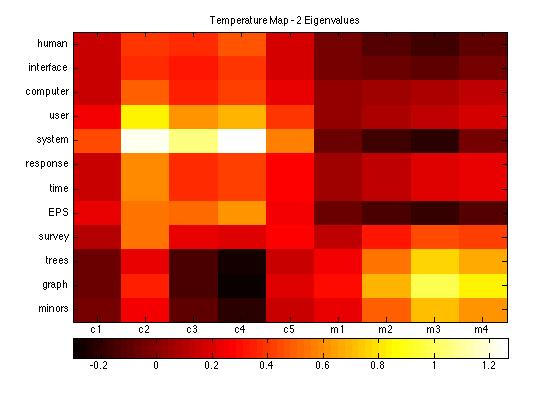}}
\end{center}
  \caption{A heatmap of an approximation to the
word-document matrix in Table 
(\ref{tab:landauerf1}) using the two largest eigenvectors.} \label{fig:heat3}
\end{figure}

In Figures (\ref{fig:heat1}) through (\ref{fig:heat3})
we have the LSA equivalent of the original Mars
photograph and its reconstructions described
above applied to the word-document matrix in Table (\ref{tab:landauerf1}).  
Figure (\ref{fig:heat1}) is a heatmap of the original matrix and to the
human eye the map appears quite sharp. There are only three colours;
black, orange and white, corresponding to whether the word
occurred zero, once or twice,  in a title. Figure  (\ref{fig:heat2}) is
analogous to  Figure (\ref{fig:yogi2}). It is ``blurry'' to the human
eye. In particular, the large black blocks in the upper right and
lower left of
Figure (\ref{fig:heat1}) have become filled with several shades of
the brown-orange colour. Figure  (\ref{fig:heat3}) is
analogous to  Figure (\ref{fig:yogi3}) and is blurrier again.

We are now in a position to explain how LSA works. 
What we would like LSA
to do, is to blur the distinction between these words in such a way
that if we searched on a word from the human-computer interaction
group, it would find all of the documents from that group and none from the 
graph theory group and vice versa. In rather more technical
language we want the
SVD and reconstruction of an approximation
of the word-document matrix 
using a smaller number of eigenvectors than 
the full word-document matrix to extract two non-overlapping
groups of words which we can identify as 
semantic categories. 
This argument regarding blurring is essentially that
put forward by \cite{Landauer2011a}
where the blurriness is compared to that which occurs in human
vision when squinting.

The reconstructions of the Mars photograph do exactly the same
thing, mathematically, as LSA. In the Mars photograph the SVD 
does not separate the photograph into eigenvectors representing
important identifiable features
such as
sky, rock, pebble, and sand which are then reassembled, feature by feature,
in the approximate reconstruction. Rather the eigenvectors are ordered, from
largest to smallest,
by the amount of variation each accounts for. Discarding
the eignevectors corresponding to the smallest 
eigenvalues removes those eigenvectors which
account for the smallest amount of variation in the original.
To the human eye this removal of small variation
is seen as the loss of fine detail. Consequently
the SVD and approximate reconstruction blurs the
distinction between sky, rock, pebble, and sand, the extent of
the blurring depends on how many eigenvalue-eigenvector pairs are
are discarded. 

\begin{table}
\begin{center}
\begin{tabular}{cccc|ccc}
Word &   &``Human''      &              &   & ``Graph'' &   \\
Rank &Original & 6 Eigen & 2 Eigen & Original & 6 Eigen & 2 Eigen \\
\hline
1 &  c4     &  c1          & c4             & m2       & m3            & m3 \\
2 &  c1     &  c4          & c2             & m3       & m4            & m4 \\
3 &  --     &  c2          & c3             & m4       & m2            & m2 \\
4 &  --     &  c3          & c5             & --       & m1            & c2 \\
5 &  --     &  m1          & c1             & --       & c2            & m1 \\
6 &  --     &  m2          & --             & --       & c3            & c5 \\
7 &  --     & --           & --             & --       & --            & -- \\
8 &  --     & --           & --             & --       & --            & -- \\
9 &  --     & --           & --             & --       & --            & -- \\
\hline
\end{tabular}
\end{center}
\caption{The relevant documents found by a search using ``human'' and``graph''
as the keywords ordered by relevance. The column ``Original''
uses  the orignal full word-document
matrix. Column ``6 Eigen'' uses the six largest eigenvalues and their 
eigenvectors. Column ``2 Eigen'' uses the two largest eigenvalues and their 
eigenvectors. }\label{tab:keywordhuman}
\end{table}
One of the key steps in LSA is trying to optimize the amount of 
blurring of the distinction
between the words that is undertaken by altering
the number of eigenvectors retained in the approximation, 
see \cite{Landauer1998} Figure (5).
In the \cite{Landauer1998} example, the goal
of an LSA
would be to remove sufficient fine detail from the word-document
matrix that the words in the two distinct document groups
become relatively indistinguishable, yet retain enough detail
that the two groups do not become merged. 

Table (\ref{tab:keywordhuman}) shows the sets of relevant documents
returned by two different keyword
searches, one word from each document group, and for each of the three
word-document matrices discussed here. From the heatmap of the
original matrix (see Figure \ref{fig:heat1}) we
can see that it is only possible
to find a relevant document if it has a colour other than black.
That is, the word must occur in the document or it will not be found.
This is exactly what is reported in the two columns labelled
``Original'' in Table (\ref{tab:keywordhuman}).

In Figure (\ref{fig:heat2}), 
the six eigenvalue approximation,
document c5 can be seen to have a very low value for ``human'',
clearly lower than any of the four document in the graph theory 
document set. In Table (\ref{tab:keywordhuman}) the search
returned six relevant documents, correctly retrieving c1-c4,
but erroneously ranking m1 and m2 as more relevant than the omitted c5.
In Figure (\ref{fig:heat3}), the two eigenvalue approximation, the extra
blurring has made documents c1-c5 more similar to each other when
searching on the keyword ``human'',
but also is now quite distinct from the m1-m4 group. 
In Table (\ref{tab:keywordhuman}) 
we can see the keyword search has correctly reported c1-c5 as
relevant while it excluded all of the graph theory documents.
In terms of optimizing the amount of blur in the word-document
matrix for the human-computer interaction group the two eigenvalue
approximation has done a better job than the six eigenvalue approximation.

The situation is reversed if we search on the word ``graph'' (second
line from the bottom in the heatmaps). In the six eigenvalue approximation
it is clear there is very little ``heat'' associated with the word ``graph''
in documents c1-c5. The results of the search in Table (\ref{tab:keywordhuman}) 
show that documents m1-m4 are the most relevant but has also included
documents c2 and c3. However, in the two eigenvalue approximation document
c2, though not relevant, is ranked higher than m1. This time,
in terms of optimizing the amount of blur in the word-document
matrix for the graph theory group, the six eigenvalue
approximation has done a better job than the two eigenvalue approximation.

The type of blurring just
discussed is like that
seen in Figures (\ref{fig:yogi2}) and (\ref{fig:yogi3})
in which the large scale features of the landscape are still visible,
fine detail is lost,
but there is new ``fine detail'' in the approximation which was not
present in the original, see, for example, the
mottled appearance of the sky.

In practice, optimizing the amount of blur is non-trivial and, as 
indiciated by the simple example, there are trade-offs to be made.
This particular example suffers from having only a small number of
words and documents, as their numbers increase so does the
ability to find an optimum amount of blur in the 
approximate word-document
matrix.

\section{Conclusions}\label{sec:conclude}

The argument put forward in this note, namely 
that LSA extracts what appears to humans to be
semantic meaning is a consequence of the blurring of the distinction
between words within a corpus. This 
idea has been advanced before
by one of the originators of LSA. But through the use of simple visualization
tools and an analogy to photographic compression, we have made this
mechanism much easier to understand for those without the necessary 
mathematical knowledge to follow the arguments in the existing literature.

\clearpage

\bibliography{LSA}

\end{document}